\documentclass[10pt,journal,final]{IEEEtran}
\usepackage{multirow} 
\usepackage{makecell}
\usepackage{xcolor,graphicx,float}
\graphicspath{{./figures/}}
\usepackage[numbers,sort&compress]{natbib}
\usepackage{amsmath}
\usepackage{amssymb}
\usepackage{bm}
\usepackage{booktabs}
\usepackage{subcaption}
\usepackage{algorithm}
\usepackage{algpseudocode}
\usepackage{setspace}
\usepackage{caption}
\usepackage{flafter}
\usepackage{placeins}
\usepackage{afterpage}

\ifCLASSOPTIONcompsoc
\usepackage[caption=false,font=normalsize,labelfont=sf,textfont=sf]{subfig}
\else
\usepackage[caption=false,font=footnotesize]{subfig}
\fi

\begin{document}

%
\title{An accelerate Prediction Strategy for Dynamic Multi-Objective Optimization}
%
%
%

\author{Ru~Lei,
        Lin~Li,
        Rustam~Stolkin,
        Bin~Feng
\thanks{Ru Lei, Lin Li and Bin Feng are with the Key Laboratory of Information Fusion Technology of Ministry of Education,
Northwestern Polytechnical University, Xi’an, Shaanxi 710072, China
(e-mail: linli@nwpu.edu.cn; leir@mail.nwpu.edu.cn; fengbin@nwpu.edu.cn;).}
\thanks{Rustam Stolkin is with the Extreme Robotics Laboratory, University of Birmingham, Birmingham B15 2TT, U.K. (e-mail: r.stolkin@bham.ac.uk).}}

%
%

\markboth{IEEE TRANSACTIONS ON EVOLUTIONARY COMPUTATION}%
{Shell \MakeLowercase{\textit{et al.}}: Bare Demo of IEEEtran.cls for IEEE Journals}
%


\maketitle

\begin{abstract}

This paper addresses the challenge of dynamic multi-objective optimization problems (DMOPs) by introducing novel approaches for accelerating prediction strategies within the evolutionary algorithm framework. Since the objectives of DMOPs evolve over time, both the Pareto optimal set (PS) and the Pareto optimal front (PF) are dynamic. To effectively track the changes in the PS and PF in both decision and objective spaces, we propose an adaptive prediction strategy that incorporates second-order derivatives to predict and adjust the algorithms search behavior. This strategy enhances the algorithm's ability to anticipate changes in the environment, allowing for more efficient population re-initialization. We evaluate the performance of the proposed method against four state-of-the-art algorithms using standard DMOPs benchmark problems. Experimental results demonstrate that the proposed approach significantly outperforms the other algorithms across most test problems.


\end{abstract}


%
\IEEEpeerreviewmaketitle

\section{Introduction}
%
%
%
%
\IEEEPARstart{N}{umerous} practical problems can be formulated as dynamic multi-objective problems (DMOPs) \cite{8Dynamic}. In simple terms, a "dynamic environment" means that the problem is constantly changing over time. For example, imagine you are trying to optimize a delivery route, but the traffic conditions, weather, and customer demands keep changing. This makes finding the best solution more challenging compared to static problems, because the conditions remain constant. As a result, the convergence process for DMOPs becomes more complex than that of static multi-objective problems. Without loss of generality, a DMOP can be formulated mathematically as: 

\begin{equation}
\begin{array}{l}
\min \; F({\bm{x},t}) = \{f_1(\bm{x},t), \cdots , f_m(\bm{x},t)\}^{T}\\
 \text{subject to}\quad \bm{x} \in \Omega = \displaystyle\prod_{i=1}^{n}\left [ {a}_{i},{b}_{i}\right ]\\
 
\end{array}
\label{Eq1}
\end{equation}
where \( F(\bm{x}, t) \) is the vector of objective functions with \( m \) time-varying objective functions \( f_j(\bm{x}, t) \),  \( j = 1, 2, \ldots, m \). \(\bm{x}\) is an \( n \)-dimensional decision variable, \(t = 1, 2, 3, \ldots \) \( T \). \( T \) denotes the time scale, and \( a_i, b_i \) are the bounds of $\bm{x_i}$ , and these constants fall within the range \((- \infty, + \infty)\) for all \( i = 1, 2, \ldots, n \).

Multi-objective evolutionary algorithms (MOEAs) are recognized as a powerful technique for solving multi-objective optimization problems (MOPs) \cite{12Dynamic,4633340}. There is a particular class of MOPs \cite{12Dynamic,MOP1,MOP2} which becomes important in various scientific and engineering applications. These MOPs are challenging, in having mutually conflicting objective functions, meaning that improving the performance of one function often comes at the expense of the others. It is impossible to optimize all such functions simultaneously. However, we can obtain a set of equilibrium solutions which is called the Pareto-optimal Set (PS) in decision variable space, which corresponds to the Pareto-optimal Front (PF) in the objective function space \cite{pspf1,pspf2}. However, in certain kinds of real-world applications, the parameters of MOPs may change over time, causing the optimal PF to evolve. These problems with changing PFs are usually called DMOPs \cite{8Dynamic}. DMOPs can be especially challenging to solve. This is partly because the objective functions are continually changing over time, and also because multiple, mutually contradictory objective functions need to be optimized at each time step. A critical challenge in solving DMOPs is that of tracking the trajectories of the changing PS and PF in various environments.


Dynamic multi-objective optimization is widely applied in dynamic scheduling \cite{scandual,Scheduling1,Scheduling2}, planning \cite{1Dynamic,2Dynamic,planning1,4Dynamic}, resource allocation \cite{resources}, and machine learning \cite{5Dynamic}. It can be difficult to adapt conventional, static MOEAs to problems where the environment changes dynamically. To address this, researchers have developed strategies to improve the convergence of solutions to DMOPs, including the diversity-based \cite{diversity2022,diversity2023,diversity2020}, memory-based \cite{memory2024,25Dynamic,memory2021}, and change prediction-based \cite{20Dynamic,prediction2022,prediction2023} methods. Diversity-based methods include mutating or re-initializing populations based on the severity of environmental changes. Memory-based methods reuse the past information by storing representative individuals to conduct future searches. Change prediction-based strategies use past search experiences to guide the evolutionary process towards future PS or PF directions, and have gained significant attention in the research community.

However, the methods above encounter several challenges when addressing certain real-world DMOPs, characterized by three main issues. First, the absence of precise guidance strategies might negatively impact subsequent optimization efforts, as they fail to provide clear directions for the algorithm to follow when changes occur. Second, approaches that utilize memory to recall past solutions often struggle to swiftly adapt to new environmental conditions, resulting in outdated or irrelevant solutions being recalled. Third, previous change prediction strategies typically focus solely on the decision space, overlooking the characteristics of the objective space. This narrow focus can lead to inaccurate convergence and suboptimal performance, as the objective space dynamics contain important additional information for effective optimization.

In this paper, we consider DMOPs with time-varying objective functions. To address these problems, we propose a novel dynamic multi-objective prediction strategy that incorporates an adaptive multi-view approach and a second-order derivative-based model. The proposed strategy can adaptively model in both the decision and objective spaces, and use these models to predict subsequent optima, thereby responding more rapidly and accurately. These two key contributions can be summarized as follows.

\begin{itemize}

\item A prediction strategy based on the second-order derivative method is proposed to reinitialize the population when a change occurs. First, this strategy applies an online k-means approach to cluster individuals, effectively capturing changes in the population distribution by continuously updating cluster centroids. Second, it leverages second-order derivative calculations from the clustering information of past populations to predict the PF's movement at different time steps. By analyzing population dynamics over time, this method accurately forecasts changes in the PF's trajectory, enabling proactive adjustments to the population. This predictive approach ensures that evolutionary algorithms can adapt efficiently to changing objectives, maintain operational effectiveness, and achieve optimal solutions in dynamic environments.

\end{itemize}

The remainder of this paper is organized as follows. Section II reviews previous literature on DMOPs. Section III presents the framework of the Second-order Derivative method and the adaptive dual-domain prediction strategy. Section IV presents experiments that evaluate the performance of our method and discusses the results. Section V provides concluding remarks and suggestions for future work.

\section{Related Work}
In this section, we first introduce the characteristics of dynamic multi-objective problems. Then we describe existing dynamic multi-objective evolutionary algorithms. Finally, we present the motivation behind our proposed evolutionary search method for solving DMOPs.

\subsection{Dynamic Multi-objective Problems Definition}
Changes in DMOPs may occur in different spaces, affecting the dynamic PS or PF accordingly \cite{8Dynamic,26Dynamic}. According to the types of change, DMOPs can be categorized into four types as follows:

Type I: PS changes, but PF does not change

Type II: Both PS and PF change

Type III: PF changes, but PS does not change

Type IV: Both PS and PF do not change

For Type IV, both the PS and PF remain static, defining them as static MOPs. Therefore, in this study, we focus only on the first three dynamic types. The main challenge in handling DMOPs lies in effectively tracking the evolving PS or PF as they change over time.

\textbf{Definition 1}$\colon$Dynamic Pareto Optimal Solution 

Following Eq. \ref{Eq1}, suppose that both $\bm{x_1}$, $\bm{x_2}$ $\in$ $\Omega$ are two individuals in the population at time \emph{t}. We consider that $\bm{x_1}$ dominates $\bm{x_2}$, represented as $\bm{x_1} \prec \bm{x_2}$, if and only if the following conditions are met:

\begin{equation}
\left\{ \begin{array}{ll}
    f_j(\bm{x}_1, t) \leq f_j(\bm{x}_2, t) & \forall j = 1, 2, \ldots, m, \\
    f_j(\bm{x}_1, t) < f_j(\bm{x}_2, t) & \exists j = 1, 2, \ldots, m.
\end{array} \right.\
\end{equation}

\textbf{Definition 2}$\colon$Dynamic Pareto Optimal Set 

The dynamic Pareto optimal set, denoted as $DP{S^*}\left( t \right)$, is a set of solutions comprising all non-dominated solutions in the decision space, if a decision vector $\bm{x}^{*}(t)$ satisfies:
\begin{equation}
DPS^{*}(t) = \{ \bm{x}^{*}(t) \mid \nexists \bm{x}(t) \prec \bm{x}^{*}(t)^{*} \}
\label{Eq2}
\end{equation}

\textbf{Definition 3}$\colon$Dynamic Pareto Optimal Front 

The dynamic optimal PF at time \emph{t}, which is denoted as $DP{F^*}\left( t \right)$, is a set formed by mapping all Pareto optimal solutions to the objective space at time $t$, which is formuled as follows:
\begin{equation}
DPF^{*}(t) = \{ f(\bm{x}(t))^{*} \mid \bm{x}^{*}(t) \in DPS^{*}(t) \}
\label{Eq3}
\end{equation}

\subsection{Related Dynamic Multi-Objective Evolutionary Algorithms}
In recent years, there has been a growing body of research literature on using evolutionary algorithms (EA) to solve DMOPs \cite{cou1,cou2,cou3,cou4}. Generally, the workflow of existing prediction methods includes two parts: change detection \cite{detection2023} and response strategies. After an environmental change, the first step is to determine if the characteristics of the problem have changed. If they have, a change response strategy, such as prediction strategy, is adopted; otherwise, a static optimization strategy is used. 
 
The DMOEAs are primarily categorized into three types: diversity maintaining methods \cite{diversity2020,diversity2022,diversity2023}, memory-based methods \cite{memory2021,25Dynamic,memory2024} and prediction methods \cite{18Dynamic,IGP,prediction2022,prediction2023}. These strategies have been developed over recent years and have been shown to be effective both theoretically and experimentally in addressing a wide range of DMOPs.

Diversity maintaining methods \cite{diversity2022,diversity2023} are designed to enhance and preserve the population's diversity upon detecting changes. Liu et al. \cite{13Dynamic} developed a diversity introduction algorithm that adaptively adjusts individuals to new environmental conditions. Chen et al. \cite{33Dynamic} have explored an evolutionary method that preserves the diversity of a collection of individuals by treating it as an additional objective in dynamic environments. Peng et al. \cite{14Dynamic} introduced a technique that uses historical data to guide the search process and increase diversity. These methods initiate individuals in the new environment through random initialization or mutation for mild changes, or through complete re-initialization for more significant changes.

Memory-based approaches \cite{31Dynamic,32Dynamic} store valuable historical information to guide future searches. Wang and Li \cite{31Dynamic} devised a multi-strategy ensemble evolutionary algorithm that uses Gaussian mutation operators and a memory-like strategy to reinitialize populations upon environmental changes. Ramsey et al. \cite{32Dynamic}, in 1993, introduced a case-based initialization for genetic algorithms, where an evolutionary mechanism updates the population by categorizing and storing past data when changes occur. However, these memory-based methods are typically best suited for scenarios where the PS remains constant or the environmental changes occur slowly. They often struggle to maintain population diversity.

Recent developments have seen a surge in integrating prediction-based strategies within evolutionary frameworks, utilizing the time-history of evolutionary steps. These strategies focus on constructing prediction models by exploiting the correlations in historical data. Iason et al. \cite{34Dynamic} have developed a novel evolutionary algorithm that incorporates forecasting techniques to address DMOPs, estimating future positions based on the time-history trajectories of optima and leveraging memory method benefits. An enhanced evolutionary prediction strategy, introduced by Anabela et al. \cite{35Dynamic}, incorporates a dynamically adjustable linear predictor. Additionally, Yang et al. \cite{36Dynamic} have devised a prediction method that segments the population into sub-populations based on reference points. When changes occur, the centers of sub-populations belonging to the same reference point are used to estimate the center sequence. Zhou et al. \cite{18Dynamic} propose a population prediction strategy (PPS), which divides the PS into two parts: center point and manifold. This method employs an autoregression model to estimate the manifold, and then combines the predicted center and manifold to generate the new population. This method only employs the previous two PS manifolds for prediction, which has small time complexity and space complexity. Cao et al. \cite{SVR} integrated a support vector regression (SVR-MOEA/D) predictor with MOEA/D to address DMOPs, in which the SVR model is trained using historical time-series data of solutions from the decision space to construct the predictor. Zhou et al. \cite{MV} focus on addressing DMOPs by using a multi-view prediction approach (MV-MOEA/D) that leverages predictions from both the decision and objective spaces. This method integrates a kernelized autoencoding model in a reproducing kernel Hilbert space (RKHS), which provides a closed-form solution and mapping strategy for multi-view predictions. Another method involving mapping in the objective space is Inverse Gaussian Process Modeling (IGP-DMOEA), proposed by Zhang et al. \cite{IGP}. This approach distinguishes itself by mapping historical optimal solutions from the objective space back to the decision space, in contrast to more conventional methods that focus solely on the decision space. This inolvement of both spaces enhances the adaptability and effectiveness of solving DMOPs.However, our approach places greater emphasis on adaptively adjusting the predicted population size within both the decision and objective spaces based on different changing states. This innovation not only fully leverages the strengths of both spaces but also dynamically optimizes the population size, leading to improved performance and robustness in solving DMOPs.

\subsection{Motivation for Conducting Second-order Derivative and Adaptive Prediction from Dual-Domain Spaces}

Dynamic changes are an inherent attribute of real-world and benchmark DMOPs, which exhibit varying patterns in the PS and PF (i.e. dynamic variations of decision space and objective space over time).
\begin{figure*}[htbp]
    \centering
    \begin{subfigure}[b]{0.8\textwidth}
        \includegraphics[width=1\linewidth,trim={70 140 150 115}, clip]{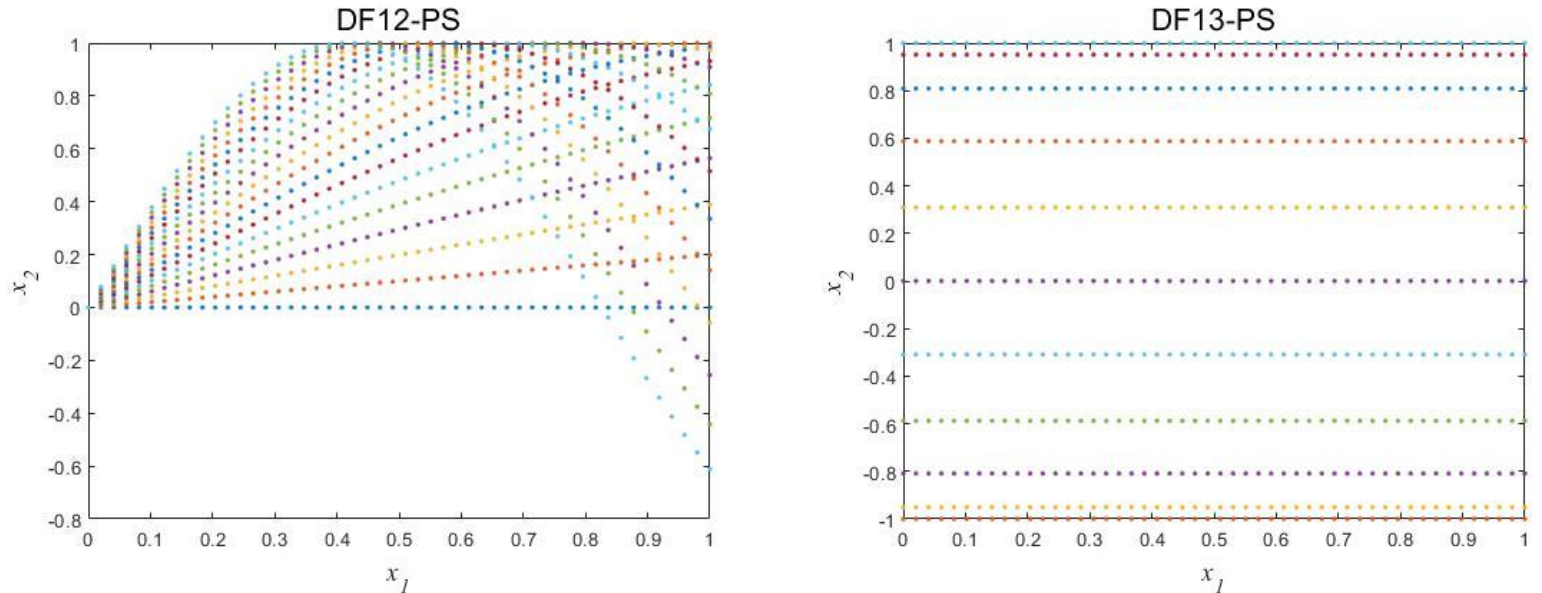}
        \caption{}
        \label{fig:subfig-a}
    \end{subfigure}
    \quad 
    \begin{subfigure}[b]{0.8\textwidth}
        \includegraphics[width=1.1\linewidth,trim={60 140 100 115}, clip]{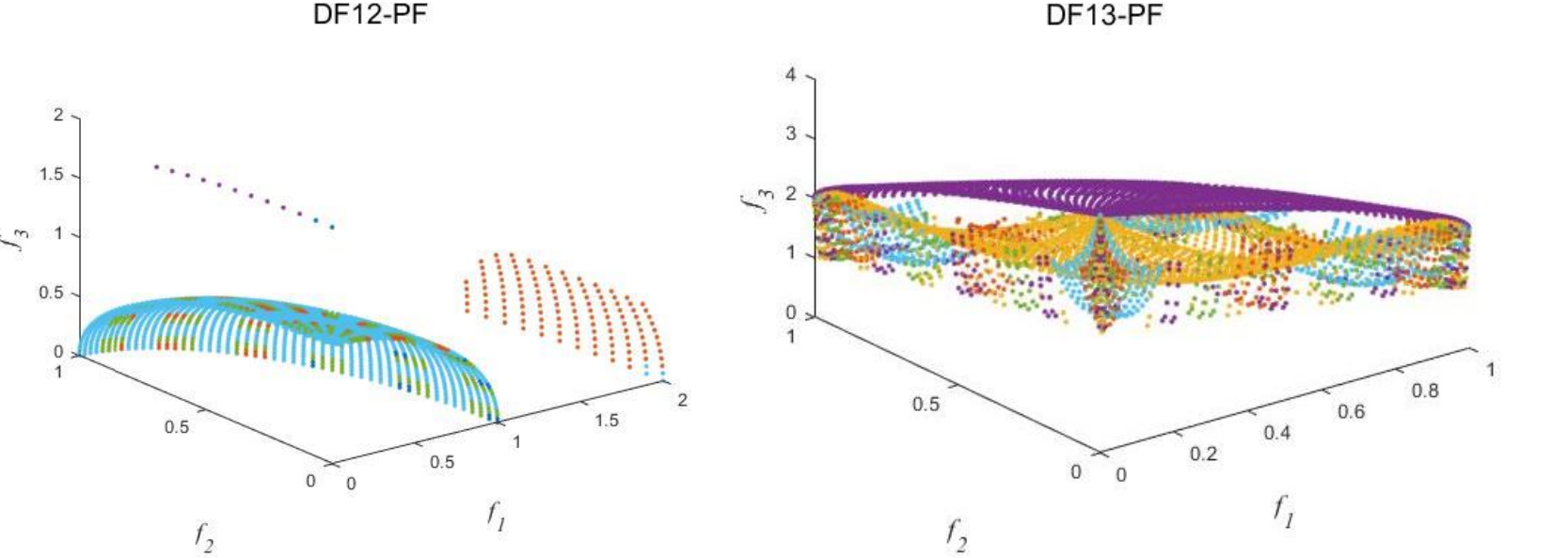}
        \caption{}
        \label{fig:subfig-b}
    \end{subfigure}
    \caption{Illustrations of the Complex Changes in PS and PF for Two Representative Type II DMOPs:(a) Changing PS of DF12 and DF13. (b) Changing PF of DF12 and DF13.}
    \label{fig:figure}
\end{figure*}

These time-varying behaviours display a diversity of patterns. For example, in the IEEE CEC2018 DMOP benchmarks \cite{CEC2018},
as shown in Fig. \ref{fig:figure}, several complex PF shapes are illustrated. As depicted in this figure, DF12 and DF13 generate both continuous and disconnected PF geometries. While the PS of DF13 is simple to track compared to the PS of DF12, the resulting PF is complex. Similar
properties can also be observed in other Type II DMOPs. Therefore, predictions made solely from the decision space or the objective space are insufficient to fully capture the diverse patterns of change. Given the unknown dynamics of DMOPs, it is crucial to establish a predictive model from both domains, to effectively handle the dynamic characteristics
and enhance convergence capabilities.

Furthermore, with complex environmental changes, the task of solving for the population at the next time-step becomes intricate and time-consuming. Thus, there is a need for efficient dynamic multi-objective optimization algorithms to enhance the convergence speed of the population. By remembering the optimal solutions from the environments at previous time-steps, it is possible to compute the direction of evolution and the changes in each evolutionary direction, ultimately obtaining the Second-order Derivative of the evolutionary direction.

In the literature, although numerous attempts have been made to develop predictive methods for dynamic multi-objective optimization, there is still relatively little work which specifically addresses complex DMOPs adaptively from both the decision and objective spaces. This paper introduces a novel method which adaptively targets the most suitable space, focusing on resource on the objective space or the decision space, depending on whichever is most useful at any given time-step, for solving the dynamic changes in both spaces. Additionally, this method integrates a Second-order Derivative prediction evolutionary search strategy with historical population analysis, aiming to effectively utilize both the positional information of historical populations, and the number of non-dominated solutions, to predict changes in the population at the next time-step.

\section{THE PROPOSED METHOD}
This section presents technical details of our proposed second-order derivative-based adaptive dual-domain prediction strategy (ADPS). ADPS works by utilizing adaptive multi-view  prediction and Second-order Derivative prediction for parallel convergence. We embed this mechanism into a decomposition-based evolutionary algorithm, MOEA/D \cite{11Dynamic}. The main processes of ADPS include dynamic detection, an adaptive dual-domain algorithm, and the second-order derivative prediction strategy. The framework of ADPS is shown in Algorithm 1.

\subsection{Dynamic Detection}
Existing methods for dynamics detection in evolutionary algorithms fall into two categories: detector-based and behavior-based detection \cite{detection,detection2}. Detector-based methods focus on reassessing specific solutions, termed detectors, to observe alterations in their functional values or their viability. On the other hand, behavior-based methods analyze the performance of the algorithm itself to detect changes dynamically.

In this paper, we employ a strategy where 10\% of the population is randomly selected to serve as detectors. At the beginning of each generation, these detectors are reassessed. If there is a discrepancy between the current and previously stored objective values of these detectors, an environmental change is considered to have occurred.

\subsection{Adaptive Dual-Domain prediction Algorithm}

Current prediction methods predominantly focus solely on the decision space, neglecting the characteristics of the objective space. In contrast, our proposed ADPS method adaptively exploits both spaces, by dynamically adjusting the search weights between the
decision versus objective spaces, depending on which space seems most informative given the recent environmental changes. Algorithm 1 presents the overall framework of the proposed method and details the specific processes of adaptation which we apply to this dual-domain paradigm.

This algorithm begins by initializing the population $N$ and the spatial weights $w_d$ and $w_o$ to 0.5, meaning that the weights are initially evenly distributed between the two spaces. In each iteration, the algorithm performs standard evolutionary operations such as reproduction and selection, while also conducting dynamic detection to identify any environmental changes.
When a change is detected, the algorithm compares the number of non-dominated solutions from the objective space, denoted as $N_{DSPF}$ with the number of non-dominated solutions from the decision space, denoted as $N_{DSPS}$. Based on this comparison, the algorithm adjusts the spatial weights $w_d$ and $w_o$ for the next iteration by increasing them with a parameter $\lambda$, The parameter $\lambda$, referred to as the weight adjustment parameter, controls the change in weights.
Subsequently, the algorithm uses a second-order derivative prediction algorithm to obtain the prediction solutions $PS_{t+1}$ and $PF_{t+1}$, as detailed in algorithm 3. Based on the weight ratios, a subset of solutions is selected from both the decision and objective spaces. These solutions are then combined to form the prediction solution set $P_{t+1}$. Finally, the algorithm employs the MOEA/D to optimize the current population.
\begin{algorithm}
\caption{Adaptive Dual-Domain Algorithm}
\begin{algorithmic}[1] 
\State \textbf{Initialization:} Initialize the population $N$ and the Spatial weight $w_d=0.5,w_o=0.5$.
\For{each iteration}
    \State  Evolutionary Search, e.g., Reproduction, Selection, etc.
    \State  Dynamic detection
    \If{environment changes}
         \If $N_{DSPF} > N_{DSPS}$
             \State $w_o = w_o + \lambda$
        \Else
            \State $w_d = w_d + \lambda$
        \EndIf
    \State Obtain the prediction solutions $PS_{t+1},PF_{t+1}$ by Second-order Derivative prediction Algorithm (Algorithm 3)
    \State Select $w_o\times N$ individuals from $PS_{t+1}$
    \State Select $w_d\times N$ individuals from $PF_{t+1}$
    \State Map $PF_{t+1}$ to the decision space  
    \EndIf
    \State Combine the prediction solutions set as $P_{t+1}$    
    \State Employ MOEA/D to optimize the current population
\EndFor
\State\textbf{End}
\end{algorithmic}
\end{algorithm}

Note that the predicted solutions in the objective space need to be mapped back to the decision space. Here, by optimizing the Euclidean distance \cite{MAP}, this method adaptively finds the optimal decision variables within the decision space, that best approximate the predicted target values in the objective space. In this way, the nonlinear relationships between the two spaces can be conveniently handled, and a set of decision vectors optimally corresponding to the target objective values is generated. The implementation process is as follows:

\begin{equation}
\operatorname{EuclideanDis} = \sqrt{\sum (Y_F - {Y_T})^2}
\label{Eq5}
\end{equation}
where $Y_F$ represents the fitness of the decision variables, $Y_T$ represents the ture objective value. The goal is to find a set of decision variables which minimizes the Euclidean distance between $Y_F$ and the given target objective value $Y_T$, which means finding the decision variables which minimize the objective function.

\subsection{Second-order Derivative Prediction Strategy}

Leveraging changes in historical data is crucial for predicting new individuals so that the population can quickly converge to the new Pareto Optimal Set in the decision space when the environment changes (and similarly the Pareto Optimal Front in the objective space). Most existing prediction-based methods rely on recording the optima from many previous environments to guide future evolution \cite{18Dynamic}. This approach requires substantial storage space and leads to information redundancy. Using only the optima from the previous time step is effectively ``zeroth order'' and does not fully capture changes in the population. In contrast, employing the optima from the previous three environments can accurately predict changes in the next time step to a second order degree.

We consider the current generation population $p{s^t}$ at time \emph{t} as well as the generation populations $p{s^{t-1}}$ and $p{s^{t-2}}$ at the previous two time steps \emph{t}-1 and \emph{t}-2.
Specifically, the difference between $p{s^{t-1}}$ and $p{s^{t-2}}$ can be used to estimate the change velocity at time \emph{t}-1, and the difference between $p{s^t}$ and $p{s^{t-1}}$ can be used to estimate the change velocity at time \emph{t}. The difference between these two velocities can then be used to estimate the rate of change of velocity, i.e., the ``change acceleration'' or second-order derivative.

\begin{figure}[!htb]
	\centering
	\includegraphics[width=1.7\linewidth,trim={200 200 100 100}, clip]{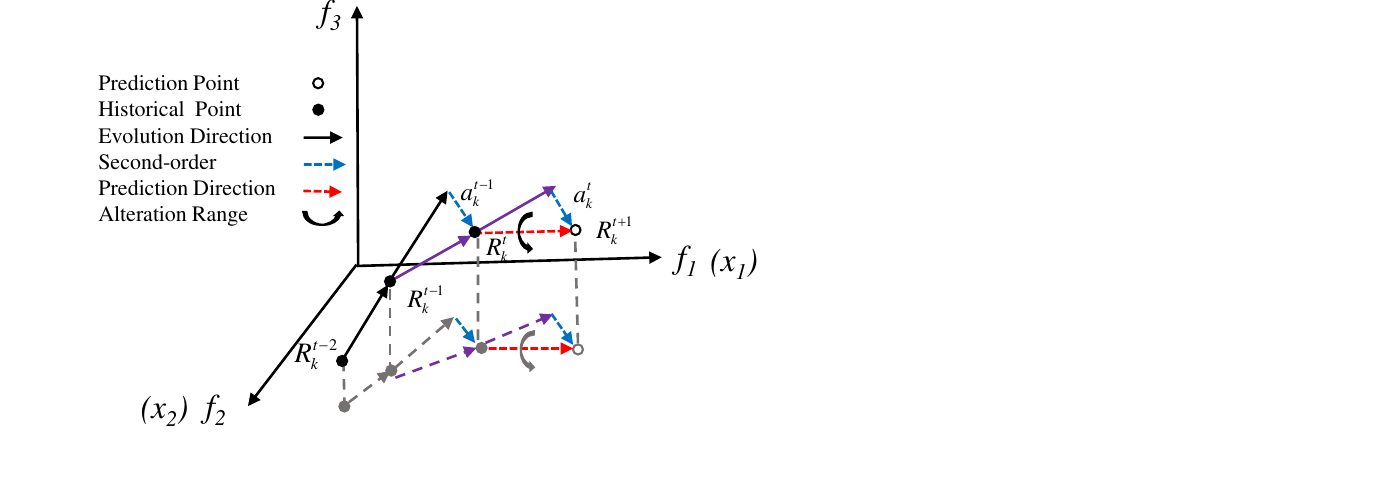}
	\caption{Second-order Derivative-based prediction strategy for DMOPs}
	\label{alpha}
\end{figure}

To fully describe the PS position at different time-steps and make accurate predictions, we employ an online k-means clustering strategy to cluster the PS at the current time \emph{t}. Subsequently, the new population is predicted based on the historical clustering centers of the PS. We begin by recording historical data: $N^t$, $N^{t-1}$, and $N^{t-2}$. Key points at times \emph{t}, \emph{t}-1, and \emph{t}-2 are then calculated and denoted as ${C^t}$, ${C^{t-1}}$, and ${C^{t-2}}$. These are then used to predict the clustering centers of the population in the new environment at the next time-step ${C^{t+1}}$.

The initial clustering centers are determined using the following technique. First, we set the number of clusters to \emph{K=M+1}. Next, the PS center point is chosen as one of the cluster centers, which represents the overall evolution direction of the population. Finally, \emph{M} boundary points are selected as additional cluster centers. These boundary points describe the location and distribution of the PF and, through their corresponding Pareto optimal solutions, determine the maximum and minimum values of the objective function, defining the edges of the PF shape. Thus, the \emph{M} boundary points, along with the center point, are used as the initial \emph{K} clustering centers, and k-means is applied to obtain the final key points for each environment. The clustering process is as shown in Algorithm \ref{alg:online_kmeans}.

\begin{algorithm}
\caption{Online K-means}
\label{alg:online_kmeans}
\begin{algorithmic}[1]
\State \textbf{Input:} Data at time $t$, number of clusters $K$, historical PS  $N^t$
\State \textbf{Otput:} Cluster centers $C^t$
\State  Set $K$ cluster centers according to the above induction
\For{each iteration}
    \For{each data point $x_i^t \in N^t$}
        \State Calculate the distance between $x_i^t$ and each cluster  center, and use Eq. (6) to find the closest center, denoted as $C^t$
        \State Update the selected center $C^t$
    \EndFor
\EndFor
\State\textbf{End}
\end{algorithmic}
\end{algorithm}

Cluster center ${C^t}$ is defined as:

\begin{equation}
{C^t} = \frac{1}{{\left| {N^t} \right|}}\sum\limits_{i = 1}^{\left| {N^t} \right|} {\bm{x}_i^t} 
\end{equation}
where $\left| {N^t} \right|$ represents the size of PS at time \emph{t}, and $\bm{x}_i^t$ represents the $i^{th}$ Pareto solution in the PS at time \emph{t}.

Let ${N^t} = \left\{ {\bm{x}_1^t,\bm{x}_2^t, \cdots ,\bm{x}_n^t} \right\}$ denote final optimal solutions at time \emph{t}, and ${C^t} = \left\{ {R_1^t,R_2^t, \cdots ,R_K^t} \right\}$ are the \emph{K} key clustering center points at time \emph{t}, then the evolution direction $\Delta {C^t}$ and the change of evolution direction ${A^t} = \left\{ {a_1^t,a_2^t, \cdots ,a_K^t} \right\}$  are calculated as:

\begin{equation}
{A^t} =  \frac{C^{t+1} - 2C^t + C^{t-1}}{\Delta t^2}\
\label{a}
\end{equation}
where \( \Delta t \) is the time interval between adjacent time steps, and the acceleration \( A^t \) is the second-order finite difference approximation of position change over time.

To effectively improve prediction accuracy for different degrees of environmental change, we use targeted methods. Here, we detect the degree of change by analyzing the relative direction between two consecutive evolutionary second-order derivatives. The difference in the angle of the relative direction is denoted as $\Delta \sin(\theta_k)$. The change in direction is obtained through the inner product of the sine differences of these two evolutionary directions, which is defined as follows:
\begin{equation}
\Delta \sin(\theta_k) = \left\langle\sin(a_k^{t + 1}) - \sin(a_k^t), \sin(a_k^t) - \sin(a_k^{t - 1})\right\rangle
\end{equation}

Note that if $\Delta \sin(\theta_k)\ge0$, it means that the two evolutionary Second-order Derivatives are in the same direction, implying that the environmental changes are not severe, and $a_k^t$ is unchanged. Conversely, if $\Delta \sin(\theta_k)<0$, the two evolutionary Second-order Derivatives are in opposite directions, indicating that the environmental changes are severe. However, relying solely on the current second-order derivative is unreliable for predicting sudden, significant changes in direction. To mitigate the impact of sudden changes, a weighted moving average method is adopted to smooth the second-order derivatives over time. Thus, the influence of historical second-order derivatives diminishes as time progresses. The final prediction can be obtained through Eq.\ref{w}.

\begin{equation}
\tilde{\bm{x}}_i^{t + 1} = \bm{x}_i^t + \Delta R_k^t  + \left\{ \begin{array}{ll}
a_k^t ,\theta \geq 0 \\
w_1 a_k^t +  \cdotp \cdotp \cdotp  + w_3 a_k^{t - 2} ,  \theta < 0
\label{w}
\end{array} \right.
\end{equation}

where $i = 1,2, \cdots ,n$, and  ${w_1} + {w_2} + {w_3} = 1$ represents the different correlations between adjacent $a^t$. Furthermore, an additional variance, $\varepsilon _i$, is required, and needs to be incorporated into Eq. \ref{w}. It can be obtained by Eq.\ref{varepsilon}, where $\bm{x}$ represents a point in the current set of solutions $N^t$, and $\bm{y}$ represents a point in the previous set of solutions $N^{t - 1}$:

\begin{equation}
0 < \varepsilon_i < \frac{1}{\max_{\bm{x} \in N^t} \left( \min_{\bm{y} \in N^{t - 1}} \left\| \bm{x} - \bm{y} \right\| \right)} \left\| \bm{x} - \bm{y} \right\|
\label{varepsilon}
\end{equation}

\begin{figure}[!htb]
	\centering
	\includegraphics[width=1.2\linewidth,trim={140 100 120 80}, clip]{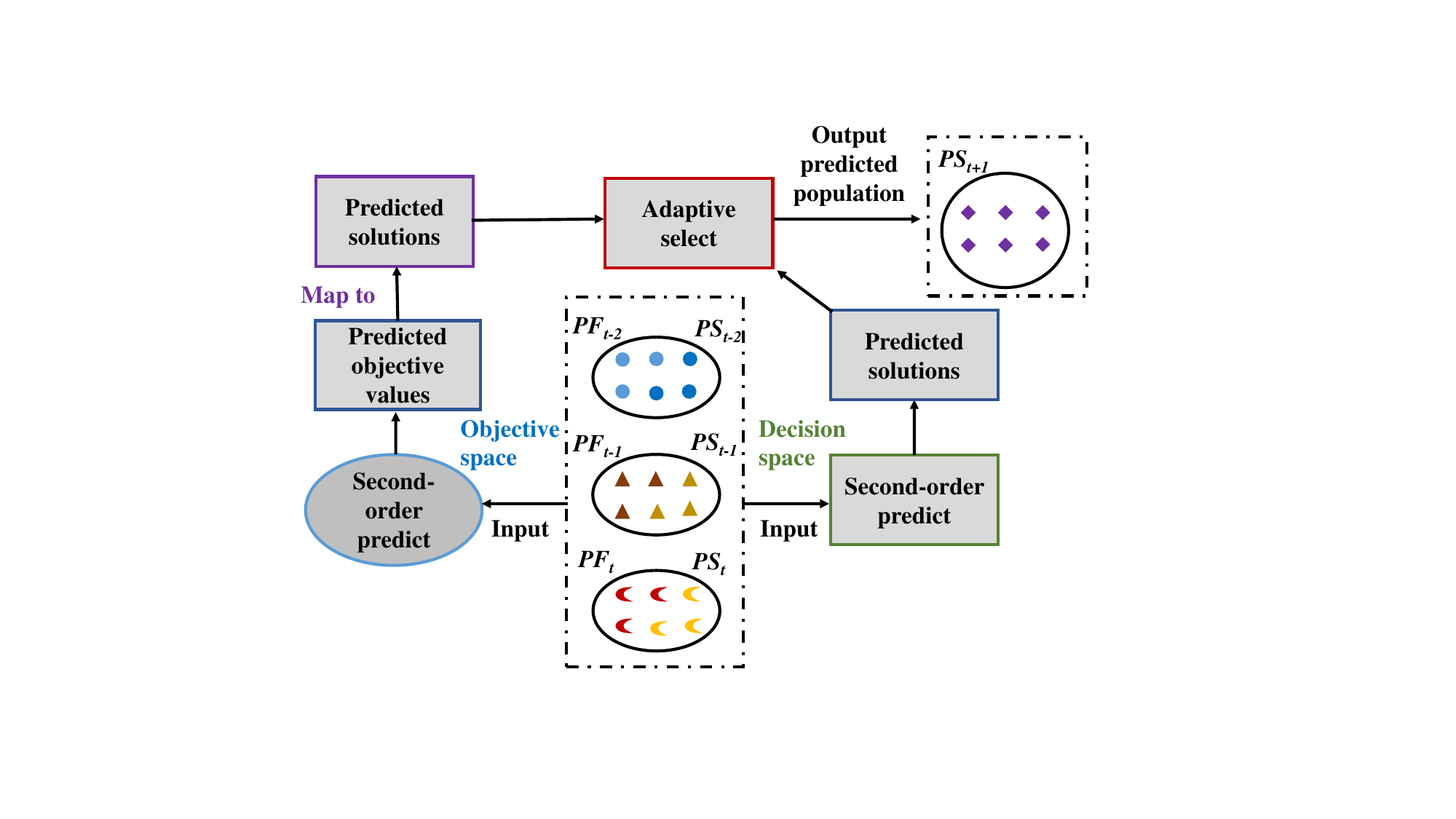} 
	\caption{Flow diagram of the proposed prediction method.}
	\label{fig:Second-order Derivative_prediction}
\end{figure}

Fig. \ref{alpha} shows the detailed prediction process. This figure provides a comprehensive representation of the second-order derivative prediction process for PF or PS. In the decision space, we calculate the prediction step size in the space \(\bm{x}_1 \text{ and } \bm{x}_2\) by considering the first two decision variables (viewed as projections in a multidimensional space). In the objective space, there are both multi-objective problems (with three or more objective functions) and two-objective problems. For two-objective problems, the approach remains consistent with the decision space. However, for for multi-objective problems, we need to combine the derivatives and step sizes from each dimension to calculate the next step prediction in the overall multidimensional space.

 Algorithm 3, using the decision space as an example, presents a Second-order Derivative-inspired prediction method. Within the framework of Algorithm 1, it utilizes clustering to estimate Second-order Derivatives as shown in Eq. (6)-(10) to predict the population at the next moment.
 
\begin{algorithm}
\caption{The Second-order Derivative Prediction Strategy }
\begin{algorithmic}[1] 
\State \textbf{Input:} Historical sub-population $P_i$
\State \textbf{Otput:} Prediction solutions in decision space $PS_{t+1}$ and objective space 
                       $PF_{t+1}$
\If{the number of \( P > 3 \)}
   \State Employ the k-means method to divide the population into \emph{k} sub-populations use Online K-means ((Algorithm 2))
   \State Calculate Second-order Derivative ${A^t}$  of each sub-populations use Eq. (7)
   \State Obtain the $PS_{t+1}$ and $PF_{t+1}$
\Else
    \State Set the prediction solutions by maintaining 90\% individuals in ${P_{t - 1}}$ and generating 10\% individuals randomly.
\EndIf 
\State Record the optimal PF and PS of the current population
\State\textbf{End}
\end{algorithmic}
\end{algorithm}

Figure \ref{fig:Second-order Derivative_prediction}  illustrates the entire proposed prediction method, detailing the processes of prediction and merging in both the objective space and the decision space. In the decision space prediction, we record the PS at three different times as historical information. After applying second-order derivative prediction and iterative evolution, we obtain the prediction solution set for the decision space. In the objective space prediction, we record the PF at three different times as historical information. Similarly, the prediction method used in the decision space is applied to obtain the objective values for the population at the next time step. Notably, predicting the objective values in the objective space requires mapping them back to the decision space. Finally, the prediction solution sets, obtained from both spaces,  are adaptively selected and merged.

\section{EXPERIMENTS}
In order to validate the effectiveness of the proposed algorithm, we conducted two sets of experiments. In the first set, we compared the performance of our algorithm against four methods, including advanced and classical approaches such as SVR-MOEA/D \cite{SVR}, PPS \cite{18Dynamic}, MV-MOEA/D \cite{MV}, and IGP-DMOEA \cite{IGP}, focusing on the accuracy of the evaluated PF compared to the true PF. In the second set, we carried out ablation studies to verify the necessity of the dual-domain perspective. These experiments utilize 14 benchmark test problems from the IEEE CEC2018 test set.

\subsection{Test Instances}
In this paper, we employ 14 benchmark functions, known as the DF test suites, to evaluate the performance of various methods. These benchmarks, first proposed by Jiang et al. \cite{CEC2018}, consist of dynamic multi-objective test problems encompassing both bi-objective and tri-objective configurations. They feature diverse dynamics, including time-dependent changes in Pareto front geometries, solution space irregularities, and discontinuities. These benchmarks are designed to simulate a broad spectrum of real-world scenarios, challenging and assessing the adaptability and efficiency of evolutionary algorithms in dynamic environments.

For all test instances, the dimension of the decision variable is 10. The parameter \( t \) is defined as:

\begin{equation}
t = \frac{1}{{n_t}}\left\lfloor \frac{\tau}{{\tau_t}} \right\rfloor
\end{equation}
where \( n_t \) is the severity of environmental change, \( \tau_t \) is the number of generations during which \( t \) remains unchanged, and \( \tau \) is the total number of generations. 

\subsection{Performance Metrics}
We introduce two widely used indicators here to evaluate the performance of the proposed algorithm. Both indicators can simultaneously measure the convergence and diversity of DMOPs approaches. This is the method commonly used for benchmarking in the research community and related literature.

1)	Invert Generation Distance (\emph{IGD}) \cite{IGDmetirc}: The IGD metric evaluates both the diversity and convergence of the obtained PF. It is calculated as follows:    (wrong)

\begin{equation}
IGD\left( {PF_t^*,{PS_t}} \right) = \frac{{\sum\nolimits_{\bm{x}_t^* \in PF_t^*} {{{\min }_{{\bm{x}_t} \in {PF_t}}}\left\| {\bm{x}_t^* - {\bm{x}_t}} \right\|} }}{{\left| {PF_t^*} \right|}}
\end{equation}
where $\left| {PF_t^*} \right|$ represents the number of individuals in $PF_t^*$. $PF_t^*$ is a set of evenly distributed points that are close to the optimal PF at time \emph{t}, and ${PF_t}$ is the true PF. The smaller the IGD is, the better the diversity and convergence of the algorithm being tested. In the experiments, 1000 and 1500 points are uniformly sampled on $PF_t^*$ of two-objective problems and three-objective problems to compute the metrics.

To measure the performance of DMOEAs, an improved IGD has been introduced \cite{18Dynamic}, defined as follows:

\begin{equation}
MIGD = \frac{\sum_{t=0}^T IGD(t)}{T}
\end{equation}
where $IGD\left( t \right)$ is the value of IGD at time \emph{t} and \emph{T} is all environments.

2)	Hypervolume Difference (\emph{HV}) \cite{HVmetirc}: The HV metric evaluates the quality of the obtained optimal PF by measuring the difference between the hypervolume of the obtained approximate PF and the true PF. HV is defined as follows:

\begin{equation}
HV_t=HV\left(PF_t\right)
\end{equation}
where $\emph{HV}(\emph{PF}_t)$ denotes the HV \cite{HV} of the set $\emph{PF}_t$. The reference point for HV calculation is $(z_j + 0.5)$, where $j = 1, \ldots, m$, and $z_j$ represents the maximum value of the $j$-th objective in the $PF_t^*$. The mean hypervolume MHV can be computed as follows:

\begin{equation}
MHV = \frac{1}{T}\sum\limits_{t = 0}^T {HV\left( t \right)} 
\end{equation}

The larger the value of MHV, the better the convergence and diversity of the corresponding method, indicating a more uniform distribution of solutions. Note that MHV cannot be calculated when the true PF is unknown.

\subsection{Parameter Settings}
We compare the performance of five dynamic multi-objective algorithms, mentioned above, against our proposed method ADPS. We choose MOEA/D as the static multi-objective optimizer for these five algorithms. The main parameters are set as follows.

1) Table \ref{tab:configuration} sets out each combination of  ${n_t}$ and ${\tau _t}$ that was tested, with 30 different environments for each instance.
\begin{table}[ht]
    \centering
    \captionsetup{
        format=plain,  
        justification=centering,  
        labelsep=newline  
    }
    \caption{ CONFIGURATIONS OF THE DYNAMIC CHANGES ON THE DMOPS}
    \label{tab:configuration}
    \begin{tabular}{c|c|c}
        \hline
        \textbf{C} & \textbf{$\tau_t$} & \textbf{$n_t$} \\
        \hline
        C1 & 5 & 5 \\
        C2 & 5 & 10 \\
        C3 & 10 & 5 \\
        C4 & 10 & 10 \\
        \hline
    \end{tabular}
\end{table}

2) The population size $P$ of MOEA/D is set as 100 for DF1-9 test functions, and 150 for DF10-14 test functions  to accommodate the increased complexity and dimensionality of the latter test cases. In order to calculate the performance metrics MIGD and MHV, 1000 points are uniformly sampled from the true PF, providing a robust basis for comparison and ensuring comprehensive coverage of the Pareto front.

3) The parameter settings, such as dimensions of decision variables, etc. are all set out in Table \ref{tab:parameters}.

\begin{table}[ht]
    \centering
    \captionsetup{
        format=plain,  
        justification=centering,  
        labelsep=newline  
    }
    \caption{PARAMETER SETTINGS}
    \label{tab:parameters}
    \begin{tabular}{c|c}
        \hline
        Parameter Variables & Parameter Values \\
        \hline
        The population size & $N = $ 100 or 150. \\
        The number of decision variables & $n = 10$. \\
        Crossover operator & $proC$ = \cite{crossover}.\\
        Mutation operator & $proM$ = \cite{mutation}. \\
        Termination criterion & $T_{max} = 30 \times \tau_t$. \\
        The number of runs & $E_{times} = 20$ times.\\
        Weight adjustment parameter & $\lambda = 0.02$.\\
        \hline
    \end{tabular}
\end{table}
4) Before the first change occurs, we set 50 generations for each algorithm.

\subsection{Experimental Results}

In this section, we first compare the PF obtained by five algorithms, by using MIGD and MHV to validate their accuracy and convergence. Next, to assess the enhancement provided by our adaptive dual-domain approach to Second-order Derivative prediction, ablation experiments are conducted. All results are derived from 20 independent trials.

\subsubsection{\emph{MIGD} and \emph{MHV} Values of Final Populations} \
The statistical results of MIGD and MHV obtained by five algorithms over 20 independent runs in dynamic environment problems, are shown in Table \ref{table1}. The mean and variance values for the 20 runs are presented. The best results for each benchmark problem are shown in bold. 
In addition, we conduct the Wilcoxon rank sum test \cite{Wilcoxon} with a 5\% significance level on the outcomes, to demonstrate the performance disparities between the baseline and the proposed algorithm. The symbols "+", "-"and "=" denote our proposed ADPS-MOEA/D being statistically significantly better, worse, or equivalent to the compared algorithm, respectively. We present a summary of the significance test results along with the MIGD and MHV scores for 14 DMOP benchmarks across four dynamic settings (amounting to 56 DF instances in total) in the bottom row of each table. The findings suggest that the proposed ADPS-MOEA/D outperforms the comparison algorithms using the same optimizer.

To explore whether the severity of dynamic changes influence to the comparative performance, we set four groups of various parameters, donated as \emph{C}1 and \emph{C}2, etc. respectively, as described above. The comparative results are shown in Table \ref{table1}. From the numerical results we can see that all the five algorithms are sensitive to the severity of change. In addition, the large values of ${n_t}$ cause the MIGD and MHV values to become smaller for most of the test problems. Therefore, all the methods demonstrate some ability to make good predictions of future promising solutions.
However, for most test problems, the MIGD of ADPS is smaller than those of its counterparts, which shows the advantages of the proposed method. ADPS has better solution distribution on DF1, DF2, DF3, DF7, DF8 and DF10. as MHV are greater than the other four methods with various ${n_t}$. This indicates that the obtained solution set is uniform and well-distributed. From the metrics of DF3, it can be observed that the results of the PPS and SVR-MOEA/D methods are slightly worse, while the MV-MOEA/D and ADPS-MOEA/D perform similarly in the MIGD metric.

Fig. \ref{PSPF} shows the true PF and PS, as well as the predicted solution sets obtained by the ADPS algorithm for four representative test problems: DF1, DF5, DF9, and DF13, compared to the solution sets from the previous time step. From the objective space PF diagrams, it is evident that for bi-objective problems, the ADPS method's predicted solutions almost match the true PF, and for tri-objective problems, they are almost always within the range of the true front, indicating that the proposed method converges well to the true PF. In the PS diagrams, the predicted populations for DF1 and DF13 resemble the true PS.

The main advantages and disadvantages of these five comparison algorithms can be summarized as follows:
\begin{itemize}

\item Our proposed ADPS method performs the best among the five algorithms, primarily due to its three key mechanisms. The first is its ability to precisely predict the direction of changes, providing the main prediction. The second is its ability to forecast the future trend of change directions, using second-order derivative change information of decision variables to guide future change directions. The third is the use of adaptive multi-view prediction, which not only analyzes historical population information from two perspectives comprehensively, but also adaptively adjusts the population distribution. 
\item  The IGP-DMOEA method uses the Inverse Gaussian Process to generate sample points in the objective space, providing accurate mapping capability, making it a well-suited comparison algorithm. 
\item  MV-MOEA/D constructs a kernel auto-encoder model, predicting from both decision space and objective space perspectives, with its performance only second to ADPS.

\clearpage 
\begin{table*}[h]
\centering
\small
 \captionsetup{
        format=plain,  
        justification=centering,  
        labelsep=newline  
    }
	\caption{Mean and Standard Deviations of MIGD Values Obtained by the Compared Algorithms with Four Dynamic Configurations}
 \label{table1}
       
    \setlength{\tabcolsep}{1mm} 
\renewcommand{\arraystretch}{1.1}  
\begin{tabular}{c|c|c|c|c|c|c}
\hline
Problem &  & SVR-MOEA/D & PPS & MV-MOEA/D & IGP-DMOEA & ADPS-MOEA/D \\
\hline
\multirow{4}{*}{DF1}  & C1 & 3.3144e-1 (6.09e-2) + & 1.0380e+0 (1.14e-1) + & 7.4855e-2 (1.66e-2) + & 8.7683e-2 (1.32e-2) + & \textbf{2.3843e-2 (3.71e-3)} \\
& C2 & 1.5104e-1 (4.76e-2) + & 5.0217e-1 (8.10e-2) + & 3.6467e-2 (8.60e-3) + & 2.8228e-2 (6.44e-3) + & \textbf{1.4502e-2 (3.09e-3)} \\
& C3 & 2.0762e-1 (7.26e-2) + & 3.5589e-1 (9.93e-2) + & 2.8146e-2 (7.22e-3) + & 7.6069e-2 (1.40e-2) + & \textbf{1.6997e-2 (3.03e-3)} \\
& C4 & 7.3563e-2 (2.59e-2) + & 1.5791e-1 (4.71e-2) + & 1.2229e-2 (2.35e-3) = & 1.4195e-2 (6.12e-3) + & \textbf{1.0864e-2 (1.94e-3)} \\
\hline

\multirow{4}{*}{DF2}  & C1 & 1.8526e-1 (3.41e-2) + & 4.1090e-1 (5.27e-2) + & 1.0121e-1 (1.12e-2) + & 1.2236e-1 (8.05e-3) + & \textbf{2.5772e-2 (6.61e-3)} \\
& C2 & 9.3251e-2 (1.81e-2) + & 1.7999e-1 (4.08e-2) + & 6.1022e-2 (8.02e-3) + & 4.8866e-2 (3.84e-3) + & \textbf{9.9069e-3 (1.17e-3)} \\
& C3 & 1.5485e-1 (2.29e-2) + & 2.7741e-1 (5.16e-2) + & 6.3470e-2 (5.57e-3) + & 9.4710e-2 (8.18e-3) + & \textbf{1.5101e-2 (3.00e-3)} \\
& C4 & 6.7104e-2 (9.29e-3) + & 1.0060e-1 (1.54e-2) + & 3.3990e-2 (5.98e-3) + & 4.9343e-2 (6.36e-3) + & \textbf{8.4918e-3 (1.59e-3)} \\
\hline

\multirow{4}{*}{DF3}  & C1 & 4.7554e-1 (1.14e-1) + & 6.5927e-2 (8.61e-3) + & 9.3100e-2 (1.07e-2) + & 6.2426e-1 (4.11e-2) + & \textbf{2.3678e-2 (1.39e-2)} \\
& C2 & 3.2128e-1 (5.28e-2) + & 2.4347e-2 (3.13e-3) + & 5.7749e-2 (1.33e-2) + & 1.4738e-1 (1.08e-1) + & \textbf{1.1070e-2 (2.55e-3)} \\
& C3 & 4.8122e-1 (8.83e-2) + & 3.3744e-2 (4.49e-3) + & 5.2773e-2 (1.37e-2) + & 5.4157e-1 (9.92e-2) + & \textbf{1.2032e-2 (1.94e-3)} \\
& C4 & 2.7616e-1 (3.32e-2) + & 1.3607e-2 (2.23e-3) + & 1.9212e-2 (5.13e-3) + & 4.0349e-2 (5.98e-2) + & \textbf{6.7041e-3 (6.53e-4)} \\
\hline

\multirow{4}{*}{DF4}  & C1 & 3.5137e-1 (3.80e-2) + & 1.8612e-1 (2.27e-2) + & 1.7212e-1 (1.35e-2) + & 7.9583e-1 (1.29e-1) + & \textbf{1.6258e-1 (1.75e-2)} \\
& C2 & 1.6762e-1 (9.17e-3) + & 1.2613e-1 (9.73e-3) + & 1.3205e-1 (5.41e-3) + & 4.7069e-1 (6.05e-2) + & \textbf{1.1818e-1 (5.54e-3)} \\
& C3 & 3.2106e-1 (2.99e-2) + & \textbf{1.1963e-1 (9.68e-3)} = & 1.3713e-1 (6.86e-3) + & 7.4572e-1 (7.99e-2) + & 1.2429e-1 (6.42e-3) \\
& C4 & 1.6221e-1 (8.25e-3) + & \textbf{1.1517e-1 (4.78e-3)} - & 1.2868e-1 (2.60e-3) + & 3.2329e-1 (1.18e-1) + & 1.2197e-1 (1.37e-3) \\
\hline

\multirow{4}{*}{DF5}  & C1 & 6.4187e-1 (1.36e-1) + & 1.3639e+0 (3.32e-1) + & \textbf{5.3606e-2 (9.12e-3)} - & 1.0979e+0 (2.14e-1) + & 1.5324e-1 (6.63e-2) \\
& C2 & 2.0271e-1 (6.57e-2) + & 6.3996e-1 (1.36e-1) + & \textbf{3.0424e-2 (5.52e-3)} - & 3.5507e-1 (1.21e-1) + & 8.8503e-2 (5.40e-2) \\
& C3 & 3.0305e-1 (4.78e-2) + & 5.0978e-1 (1.96e-1) + & \textbf{2.2964e-2 (2.76e-3)} - & 7.8335e-1 (2.31e-1) + & 1.2186e-1 (7.21e-2) \\
& C4 & 8.5015e-2 (1.26e-2) + & 1.4838e-1 (3.50e-2) + & \textbf{1.2825e-2 (1.88e-3)} - & 1.2314e-1 (3.21e-2) + & 5.4919e-2 (1.56e-2) \\
\hline

\multirow{4}{*}{DF6}  & C1 & 1.7272e+0 (1.80e-1) + & 2.4613e+0 (1.52e-1) + & \textbf{5.2719e-1 (3.38e-2)} - & 2.2755e+0 (9.32e-2) + & 1.5386e+0 (1.30e-1) \\
& C2 & 1.1026e+0 (1.13e-1) = & 1.8656e+0 (1.23e-1) + & \textbf{4.8644e-1 (2.80e-2)} - & 1.8045e+0 (1.33e-1) + & 1.1431e+0 (1.41e-1) \\
& C3 & 1.9604e+0 (1.71e-1) + & 1.9521e+0 (3.19e-1) + & \textbf{5.7317e-1 (4.23e-2)} - & 2.4768e+0 (1.52e-1) + & 1.2113e+0 (1.06e-1) \\
& C4 & 1.0007e+0 (2.22e-1) = & 1.2964e+0 (2.05e-1) + & \textbf{5.4213e-1 (3.34e-2)} - & 2.0154e+0 (1.21e-1) + & 1.0011e+0 (1.46e-1) \\
\hline

\multirow{4}{*}{DF7}  & C1 & 2.5162e-1 (8.15e-2) + & 1.4310e-1 (6.01e-2) + & 1.1678e-1 (4.20e-2) + & 1.4250e-1 (1.75e-2) + & \textbf{1.3008e-2 (6.14e-4)} \\
& C2 & 8.3902e-2 (2.31e-2) + & 4.9720e-2 (2.22e-2) + & 3.1228e-2 (1.66e-2) + & 1.8487e-2 (7.46e-4) + & \textbf{1.1546e-2 (2.29e-4)} \\
& C3 & 6.5542e-2 (1.74e-2) + & 4.1323e-2 (1.91e-2) + & 2.4073e-2 (4.31e-3) + & 1.5983e-1 (2.74e-2) + & \textbf{9.5341e-3 (3.89e-4)} \\
& C4 & 3.3255e-2 (7.01e-3) + & 2.7826e-2 (1.03e-2) + & 1.5023e-2 (2.69e-3) + & 1.6513e-2 (1.06e-3) + & \textbf{8.4630e-3 (1.76e-4)} \\
\hline
\multirow{4}{*}{DF8}  & C1 & 2.2994e-2 (2.63e-3) + & 9.6557e-3 (2.79e-4) + & 8.8864e-3 (3.59e-4) + & 9.6432e-3 (4.20e-4) + & \textbf{7.8713e-3 (3.79e-4)} \\
& C2 & 1.6890e-2 (1.07e-3) + & 9.2834e-3 (2.28e-4) + & 8.5336e-3 (3.73e-4) + & 9.2800e-3 (1.79e-4) + & \textbf{7.4589e-3 (3.79e-4)} \\
& C3 & 2.5122e-2 (2.48e-3) + & 1.1987e-2 (2.06e-4) + & 1.0771e-2 (1.07e-3) + & 1.2426e-2 (2.47e-4) + & \textbf{9.1916e-3 (5.31e-4)} \\
& C4 & 1.9039e-2 (1.52e-3) + & 1.1179e-2 (2.93e-4) + & 1.0228e-2 (1.39e-3) + & 1.1861e-2 (2.70e-4) + & \textbf{7.6506e-3 (5.01e-4)} \\
\hline

\multirow{4}{*}{DF9} & C1 & 7.7610e-1 (1.53e-1) + & 1.5390e+0 (2.07e-1) + & \textbf{7.1971e-2 (1.01e-2)} - & 8.2503e-1 (5.55e-2) + & 1.2355e-1 (1.92e-2) \\
& C2 & \textbf{3.5034e-2 (3.95e-3)} - & 9.4815e-1 (1.50e-1) + & 3.3888e-1 (1.14e-1) + & 3.2585e-1 (4.94e-2) + & 5.8663e-2 (1.09e-2) \\
& C3 & 6.5369e-1 (1.33e-1) + & 1.1614e+0 (3.98e-1) + & \textbf{4.3751e-2 (3.53e-3)} - & 7.0464e-1 (5.86e-2) + & 5.7526e-2 (7.00e-3) \\
& C4 & 3.2334e-1 (8.04e-2) + & 4.0192e-1 (1.39e-1) + & \textbf{2.2743e-2 (1.14e-3)} - & 1.8795e-1 (1.48e-2) + & 2.9801e-2 (3.60e-3) \\
\hline

\multirow{4}{*}{DF10}  & C1 & 6.3120e-1 (1.55e-1) + & 4.2983e-1 (6.96e-2) + & 1.9120e-1 (2.87e-2) + & 1.2653e+0 (2.32e-1) + & \textbf{1.6815e-1 (2.44e-2)} \\
& C2 & 2.2178e-1 (3.09e-2) + & 2.8673e-1 (4.30e-2) + & 1.5585e-1 (1.78e-2) = & 5.4133e-1 (2.44e-1) + & \textbf{1.4845e-1 (5.12e-2)} \\
& C3 & 4.1771e-1 (5.93e-2) + & 3.1460e-1 (4.22e-2) + & 1.8092e-1 (2.64e-2) + & 1.3424e+0 (1.51e-1) + & \textbf{1.3002e-1 (1.36e-2)} \\
& C4 & 2.0285e-1 (1.15e-2) + & 2.3579e-1 (2.87e-2) + & 1.4907e-1 (2.02e-2) + & 8.2863e-1 (4.92e-1) + & \textbf{1.1813e-1 (1.21e-2)} \\
\hline
\multirow{4}{*}{DF11}  & C1 & 2.3408e-1 (5.82e-3) + & 2.2005e-1 (2.83e-3) + & 2.0906e-1 (1.46e-3) + & 2.3112e-1 (8.44e-3) + & \textbf{2.0693e-1 (2.08e-3)} \\
& C2 & 2.1110e-1 (1.70e-3) + & 2.1339e-1 (1.85e-3) + & 2.0599e-1 (1.27e-3) + & 2.1125e-1 (4.37e-3) + & \textbf{2.0496e-1 (1.26e-3)} \\
& C3 & 2.3822e-1 (5.28e-3) + & 2.1854e-1 (1.93e-3) + & 2.1128e-1 (1.81e-3) = & 2.1847e-1 (6.90e-3) + & \textbf{2.1070e-1 (1.45e-3)} \\
& C4 & 2.1707e-1 (3.11e-3) + & 2.1424e-1 (1.34e-3) + & 2.0923e-1 (1.22e-3) = & \textbf{2.0744e-1 (3.73e-3)} - & 2.0893e-1 (1.49e-3) \\
\hline

\multirow{4}{*}{DF12} & C1 & 3.2413e-1 (5.15e-2) + & 3.3510e-1 (4.39e-2) + & 1.2770e-1 (1.42e-2) = & 5.0650e-1 (2.48e-2) + & \textbf{1.2556e-1 (8.90e-3)} \\
& C2 & 1.6885e-1 (7.82e-3) + & 2.2443e-1 (3.26e-2) + & \textbf{1.0605e-1 (9.35e-3) -} & 3.0078e-1 (7.54e-2) + & 1.2273e-1 (8.10e-3) \\
& C3 & 2.6156e-1 (2.73e-2) + & 2.6911e-1 (3.55e-2) + & 1.2616e-1 (1.66e-2) + & 4.5090e-1 (3.85e-2) + & \textbf{1.1752e-1 (1.02e-2)} \\
& C4 & 1.7142e-1 (4.89e-3) + & 2.0027e-1 (2.64e-2) + & 1.2066e-1 (8.38e-3) = & 1.3715e-1 (2.83e-2) + & \textbf{1.1122e-1 (4.42e-3) }\\

\hline
\multirow{4}{*}{DF13}  & C1 & 6.3528e-1 (1.36e-1) + & 5.4481e-1 (1.02e-1) + & \textbf{1.9338e-1 (1.10e-2) -} & 1.1068e+0 (4.44e-1) + & 2.4353e-1 (3.55e-2) \\
& C2 & 2.7746e-1 (1.64e-2) + & 3.6595e-1 (4.87e-2) + & 1.9108e-1 (9.87e-3) = & 7.0542e-1 (3.79e-1) + & \textbf{1.8895e-1 (2.16e-2) }\\
& C3 & 5.0658e-1 (7.98e-2) + & 3.9995e-1 (6.06e-2) + & \textbf{1.8525e-1 (1.59e-2) -} & 8.7984e-1 (4.32e-1) + & 2.0959e-1 (2.53e-2) \\
& C4 & 2.6752e-1 (1.63e-2) + & 2.8546e-1 (2.70e-2) + & \textbf{1.7991e-1 (9.83e-3) - }& 5.4950e-1 (2.38e-1) + & 1.9834e-1 (9.46e-3) \\
\hline
\multirow{4}{*}{DF14}  & C1 & 3.7787e-1 (6.94e-2) + & 4.9833e-1 (1.28e-1) + & \textbf{1.5494e-1 (6.19e-3)} - & 4.1884e-1 (4.63e-2) + & 2.6739e-1 (5.18e-2) \\
& C2 & 2.3598e-1 (5.15e-2) = & 3.3047e-1 (8.30e-2) + & \textbf{1.5727e-1 (9.20e-3)} - & 2.5191e-1 (5.03e-2) = & 2.2803e-1 (3.03e-2) \\
& C3 & 3.1065e-1 (4.18e-2) + & 3.6818e-1 (8.71e-2) + & \textbf{1.5343e-1 (5.69e-3)} - & 4.0536e-1 (3.49e-2) + & 2.3703e-1 (6.71e-2) \\
& C4 & 1.7887e-1 (1.87e-2) = & 2.4171e-1 (3.22e-2) + & 1.7854e-1 (1.23e-2) = & 1.6401e-1 (2.71e-2) + & \textbf{1.7841e-1 (1.32e-2)} \\
\hline

+/-/= & & 51/1/4 & 54/1/1 & 29/18/9 & 54/1/1 & ——\\
\hline
best/all & & 1/56 & 2/56 & 18/56 & 1/56 & 34/56 \\
\hline
\end{tabular}

\end{table*}
\clearpage 

\begin{figure}[htbp]
    \centering
    \begin{subfigure}[b]{0.45\textwidth}
        \includegraphics[width=1.4\linewidth,trim={150 100 120 40}, clip]{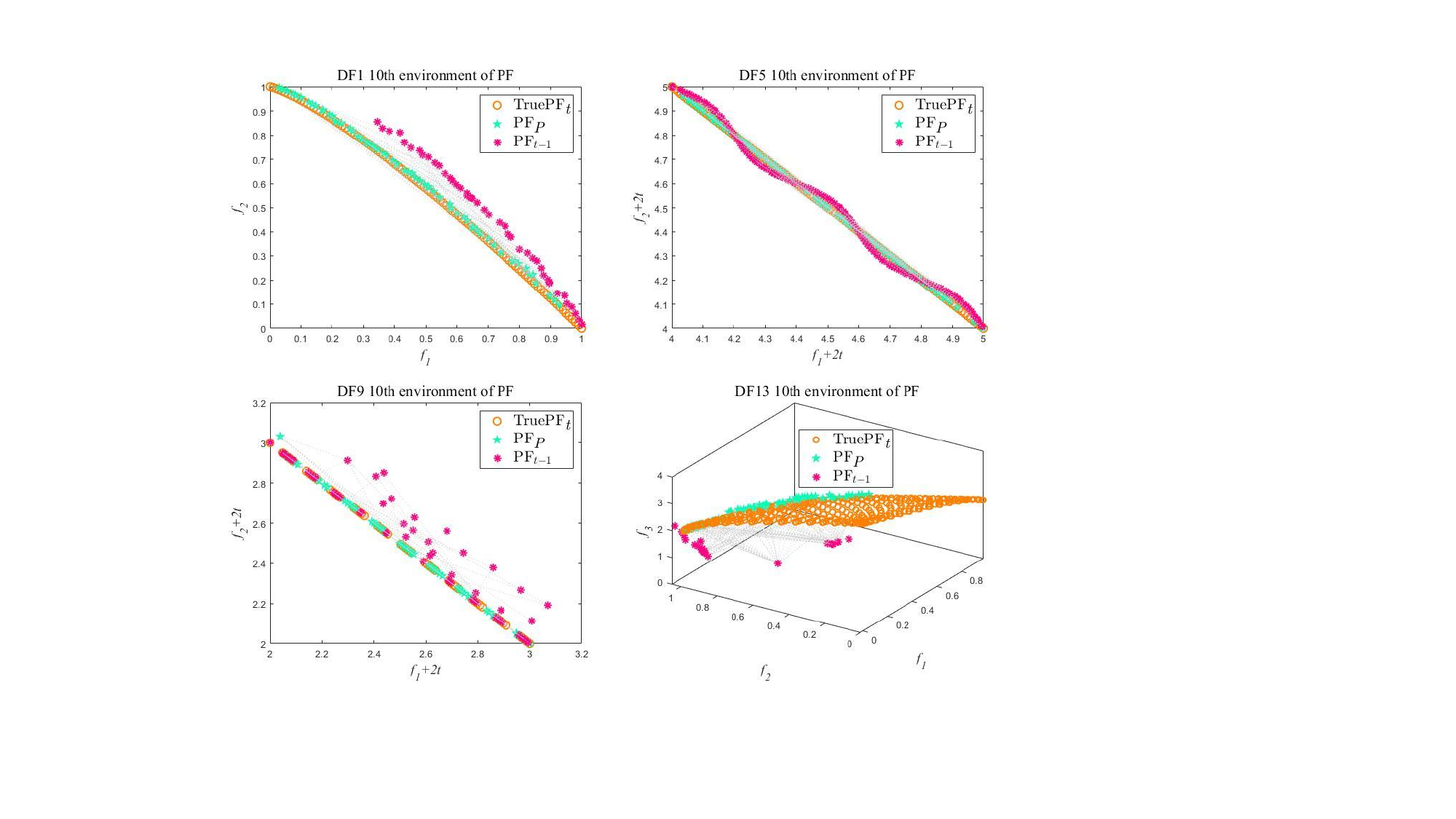}
        \caption{}
        \label{fig:pspf-a}
    \end{subfigure}
    \begin{subfigure}[b]{0.45\textwidth}
        \includegraphics[width=1.4\linewidth,trim={140 100 120 40}, clip]{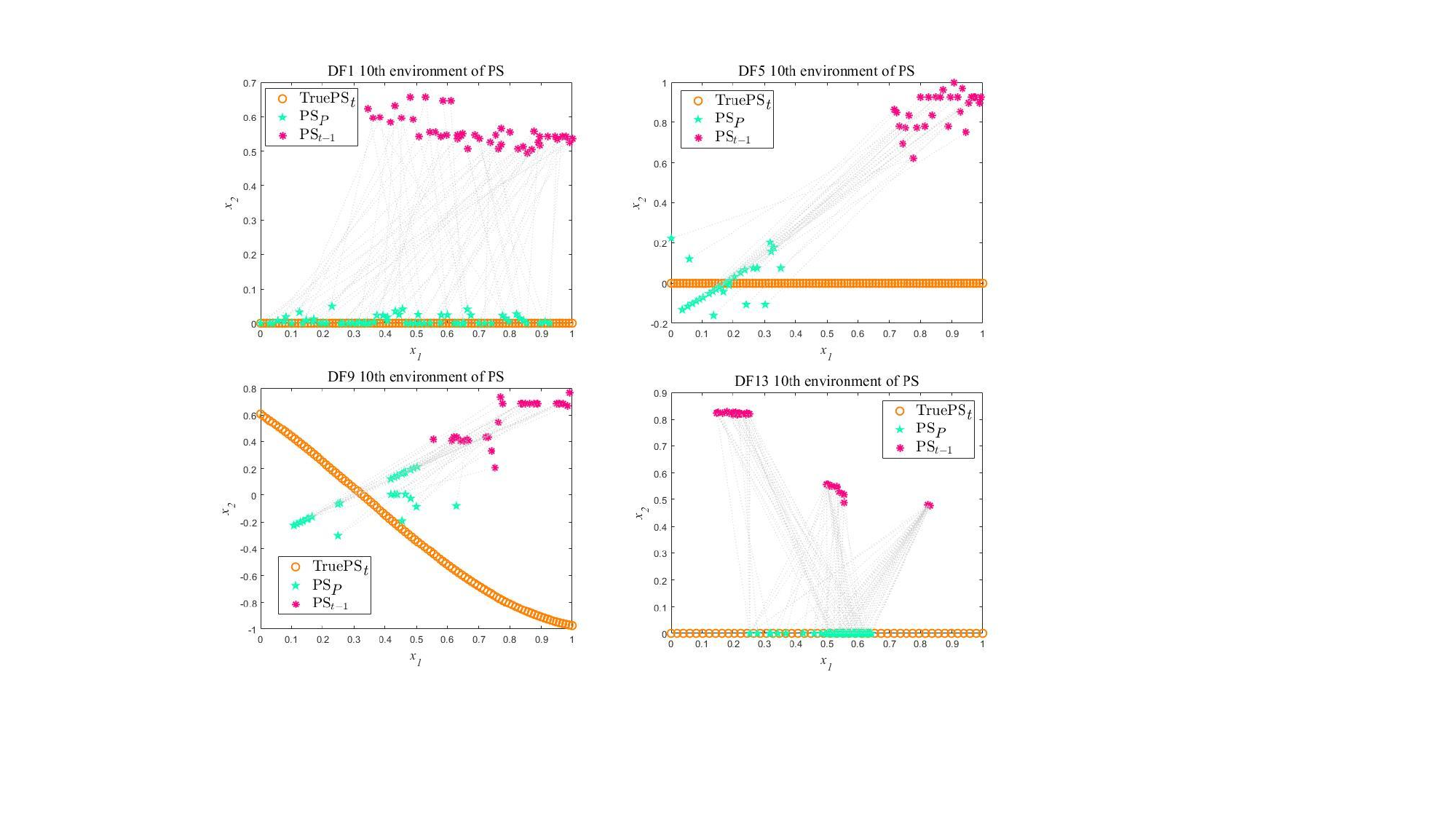}
        \caption{}
        \label{fig:pspf-b}
    \end{subfigure}
    \caption{Illustration of the ADPS strategy in DF1, DF5, DF9 and DF13 with dynamic configuration \emph{C3}, where the red markers represent the objective values of \emph{$PF_{t-1}$}, \emph{$PS_{t-1}$}. The green markers represent the two-dimensional decision variable values of \emph{$PF_P$}, \emph{$PS_P$}, evaluated in environment \emph{t}. The dotted lines indicate the direction of evolution from each solution at time \emph{t-1} to time \emph{t}, and the orange markers denote the true PF, PS of environment \emph{t}. (a) Illustration of PF. (b) Illustration of PS. }    
    \label{PSPF}
\end{figure}

\item  On most test problems, PPS and SVR-MOEA/D perform the worst. PPS uses a linear regression model based on previous individuals to predict changes in the new environment, while SVR-MOEA/D trains an SVR model using historical time-series solutions. However, both of these mechanisms operate only within the decision space, which may lead to inaccurate predictions of change types.
\end{itemize}


\begin{figure}[!htbp]
	\centering
	\includegraphics[width=0.42\textwidth]{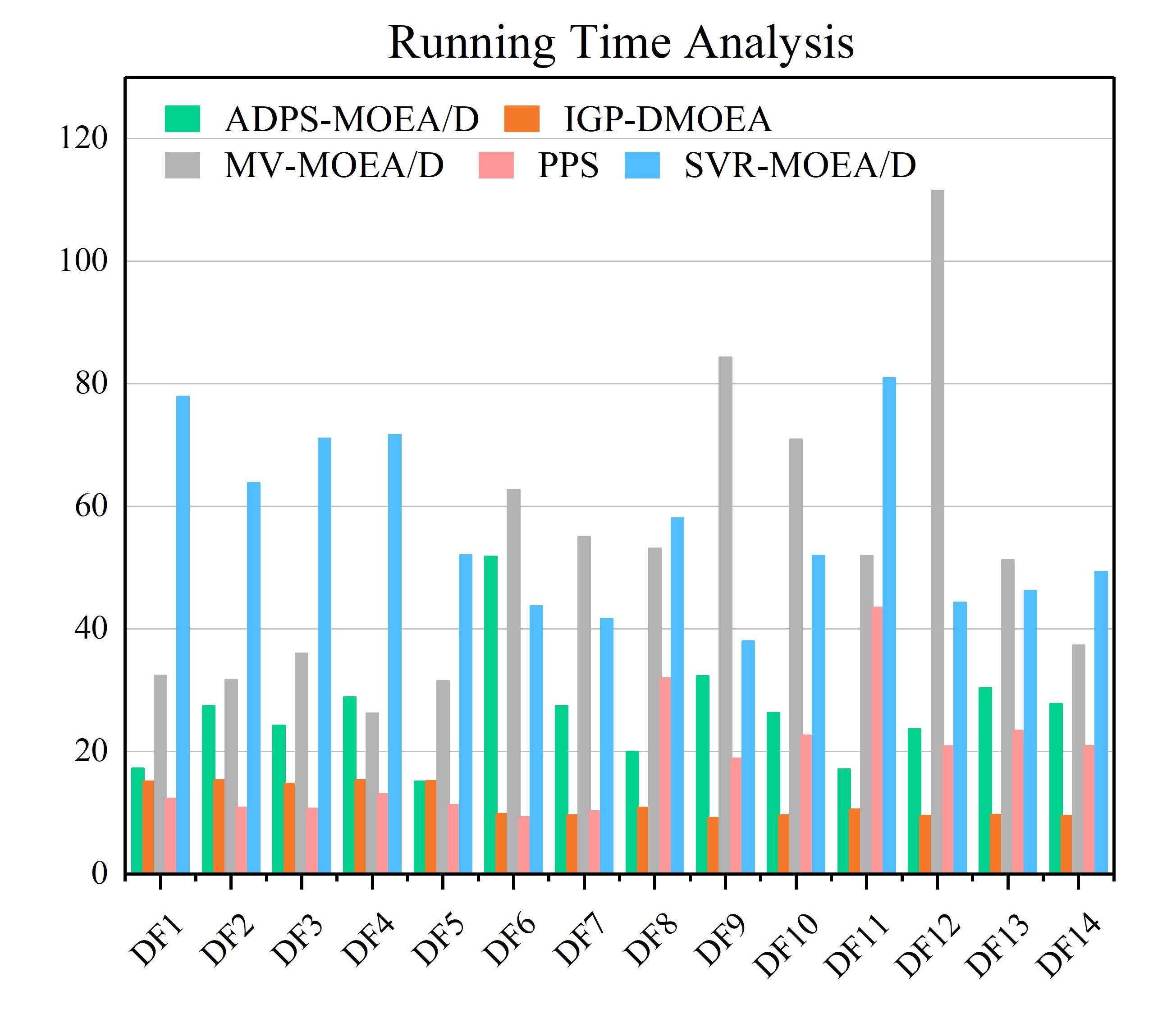} 
	\caption{The average running times of ADPS-MOEA/D and its four competitors on DF test problems with dynamic configuration C3.}
	\label{time}
\end{figure}

Clearly, the prediction-based DMOEA inevitably incurs additional computational overhead for training the prediction model. Therefore, to further investigate the actual runtime of ADPS-MOEA/D and its four competitors, a histogram of their average runtime over 20 runs on DF test problems is plotted in Fig. \ref{time}.

It is evident that PPS generally runs faster than its competitors in most cases because PPS does not involve model training and mapping issues. However, PPS also shows the worst MIGD and MHV overall performance. SVR-MOEA/D, consumes the most runtime, significantly more than its competitors. This is likely due to the extensive resources required to train the SVR model. IGP-DMOEA, involving training of the Inverse Gaussian Process model, takes less time than the de-noising autoencoder of MV-MOEA/D, but has lower prediction accuracy than the latter. Although embedding learning models such as machine learning, transfer learning, and deep learning into DMOEA can enhance performance, it inevitably adds extra overhead to the learning models. Therefore, it is essential to consider the associated costs when designing DMOEA prediction models. Notably, our proposed ADPS-MOEA/D runs relatively quickly on the DF test problems. Although ADPS involves clustering and model training processes for mapping, our Second-order Derivative prediction algorithm can expedite convergence, showing the strongest overall performance, which significantly enhances its optimization performance on most tested DMOPs.

\subsubsection {Ablation Study}\
In this section, we conducted an ablation experiment to illustrate the effectiveness of the adaptive dual-domain prediction strategy proposed in our Second-order Derivative prediction algorithm. In this experiment, we modified the original ADPS by isolating the prediction mechanisms into two variants: ADPS-I, which makes predictions solely in the decision space, and ADPS-II, which operates exclusively in the objective space. This segmentation was intended to evaluate the individual contributions of predicting within each space separately, thereby substantiating the utility of integrating both spaces in the adaptive dual-domain approach. The goal was to demonstrate that the combined strategy significantly enhances prediction accuracy and robustness, compared to using any single-space prediction method alone.

\begin{table}[H]
\centering
\setlength{\tabcolsep}{1mm} 
 \captionsetup{
        format=plain,  
        justification=centering,  
        labelsep=newline  
    }
 \caption{COMPARISON OF RESULTS OF ADPS-DMOEA VERSUS ITS TWO VARIANTS ON DF TEST PROBLEMS}
 \label{table ablation}   
\begin{tabular}{c|c|c|c}
\hline
MIGD & +/-/= & MHV & +/-/=  \\
\hline
 ADPS vs ADPS-I& 44/3/9 &  ADPS vs ADPS-II & 46/2/8  \\
 ADPS vs ADPS-I& 24/7/25 &  ADPS vs ADPS-II & 11/8/27   \\
\hline
\end{tabular}
\end{table}

Table \ref{table ablation} illustrates  the results from an ablation study designed to evaluate the efficacy of ADPS-MOEA/D and its two single-domain variants when faced with DF test problems. The symbols "+",”-”and "=" denote our proposed ADPS-MOEA/D being statistically significantly better, worse, or equivalent to the ablation algorithms, respectively. This analysis clearly demonstrates the superior performance of the dual-domain ADPS-MOEA/D, which operates in both the decision and objective spaces, emphasizing the value of our adaptive dual-domain approach. The adaptive approach allows the algorithm to effectively combine insights from both decision and objective spaces, while dynamically weighting the contributions from each domain in response to dynamically changing environments. This achieves significantly more reliable and superior performance outcomes than either of the single-domain variants. This capability is particularly critical in dynamic environments where conditions and requirements can shift unpredictably.

In the MHV category, ADPS-MOEA/D not only substantially outperforms ADPS-II, but also maintains a strong performance improvement over ADPS-I. The consistently superior performance, across different metrics and test scenarios, confirms the substantial benefits of the adaptive dual-domain mechanism in real-world applications. The adaptability is key to the algorithm's ability to fine-tune its operations in response to continual environmental changes, thereby maintaining high levels of accuracy and robustness. The success of ADPS-MOEA/D underscores the necessity of employing adaptive strategies in complex, dynamically changing environments, proving that flexibility and responsiveness are indispensable attributes for predictive algorithms.

In conclusion, the ablation study outlined in this section clearly demonstrates the significant advantage of employing an adaptive dual-domain strategy within the ADPS-MOEA/D framework. The experimental results illustrate that the full ADPS strategy, which dynamically integrates both decision and objective spaces, significantly outperforms its variants that operate within a single domain. This adaptability allows ADPS to effectively respond to a wide range of environmental changes, maintaining high levels of prediction accuracy and robustness. Particularly, the adaptive approach shows its strength in managing complex dynamics, where conventional methods might fail to adapt quickly. The success of the ADPS framework in these experiments underscores the importance of adaptability in predictive modeling, especially in environments that are subject to frequent and unpredictable changes.

\section{CONCLUSION}

In this study, we introduced a novel, Second-order Derivative-based, adaptive dual-domain prediction strategy for solving DMOPs. The paper contributes significantly in two main areas. Firstly, we developed an adaptive dual-domain strategy that simultaneously considers changes in both the PS and PF, thereby enhancing the accuracy and responsiveness of environmental change prediction. Secondly, we introduced a Second-order Derivative-based prediction strategy, which effectively re-initializes the population when changes are detected. By using historical data to precisely forecast new population dynamics, our ADPS method ensures that the population is better adapted to the new environment, thereby improving optimization performance. 

To evaluate the efficacy of our proposed method, we conducted two distinct sets of experiments. The first set compared our ADPS method against four other state-of-the-art techniques from recent literature, assessing the optima from all five algorithms to explore the diversity and convergence properties of the solutions. The second set of experiments were ablation studies, to verify the effectiveness of our adaptive dual-domain approach. This was achieved by isolating the impact of each component of our method, and confirming their individual and collective contributions to overall performance. The experimental results demonstrate our ADPS algorithm's advantages in terms of improved convergence and distribution in dynamic environments.

Nevertheless, if an environment changes extremely rapidly by large amounts, the proposed ADPS method might still fail to predict the change direction effectively. Additionally, the training models used for mapping also require optimization to enhance the accuracy of predictions while reducing computational costs. Future research should focus on developing innovative approaches that offer more precise predictions for managing larger and more diverse environmental changes. This could perhaps involve integrating more sophisticated machine learning models, and improving the efficiency of the prediction algorithms to handle real-time updates more effectively.



\ifCLASSOPTIONcaptionsoff
  \newpage
\fi

\bibliographystyle{IEEEtran}
\bibliography{refer}

\end{document}